\documentclass{article}

\usepackage{arxiv}

\usepackage[utf8]{inputenc} 
\usepackage[T1]{fontenc}    
\usepackage{hyperref}       
\usepackage{url}            
\usepackage{booktabs}       
\usepackage{amsfonts}       
\usepackage{nicefrac}       
\usepackage{microtype}      
\usepackage{lipsum}
\usepackage{graphicx}

\usepackage[utf8]{inputenc} 
\usepackage[T1]{fontenc}    
\usepackage{hyperref}       
\usepackage{url}            
\usepackage{booktabs}       
\usepackage{amsfonts}       
\usepackage{nicefrac}       
\usepackage{microtype}      
\usepackage{lipsum}
\usepackage{xcolor}
\usepackage{algorithm}
\usepackage{algorithmic}
\usepackage{amsmath}

\DeclareMathOperator*{\argmin}{arg\,min}

\DeclareMathOperator*{\minmax}{min,max}

\title{Chosen methods of improving small object recognition with weak recognizable features}

\author{
  Magdalena Stacho\'n\\
  Institute of Computer Science\\
  AGH University of Science and Technology\\
  Cracow, Poland \\
  \texttt{stachon@agh.edu.pl} \\
   \And
 Marcin Pietro\'n \\
  Institute of Electronics\\
  AGH University of Science and Technology\\
  Cracow, Poland \\
  \texttt{pietron@agh.edu.pl} \\
}

\begin{document}
\maketitle

\begin{abstract}
Many object detection models struggle with several problematic aspects of small object detection including the low number of samples, lack of diversity and low features representation. Taking into account that GANs belong to generative models class, their initial objective is to learn to mimic any data distribution. Using the proper GAN model would enable augmenting low precision data increasing their amount and diversity. This solution could potentially result in improved object detection results. Additionally, incorporating GAN-based architecture inside deep learning model can increase accuracy of small objects recognition. In this work the GAN-based method with augmentation is presented to improve small object detection on VOC Pascal dataset. The method is compared with different popular augmentation strategies like object rotations, shifts etc. The experiments are based on FasterRCNN model.
\end{abstract}


\section{Introduction}

Computer vision relays deeply on object detection including such domains as self-driving cars, face recognition, optical character recognition or medical image analysis. Over the past years, great progress has been made with the appearance of deep convolutional neural networks or recently transformer based models \cite{}. The first, based on regional nomination methods, such as the R-CNN model family \cite{rcnn-explain}, the other - one-stage detector, which enables real-time object detection with methods such as YOLO \cite{yolo-explain} or SSD \cite{ssd-explain} architectures. For those models very impressive results have been achieved for high resolution, clear objects, however, this process does not apply to very small objects. The deep learning models enable creating low level features, which afterwards are combined into some higher level features that the network aims to detect. 
Due to significant image resolution reduction, small object features, extracted on the first layers disappear in the next layers and are not affected by detection and classification scopes. Their poor quality appearance makes impossible to distinguish them from the other categories. Small object accurate detection is crucial for many disciplines and determines their credibility and effective usage. Small traffic signs and objects detection influence the self-driving car safety rules. In medical image diagnosis, a few pixel size tumor detection or chromosome recognition enables early treatment. To make full use of satellite image inspection many small objects need a precise annotation. Taking some of those examples into account, the small objects are present in every computer vision aspect, and they should be treated with special attention, as they constitute one of the weakest parts of current object detection mechanisms. In this work few approaches were tested and compared how they can help in improving detection of small objects. First the most popular methods were taken based on augmentation techniques. In the next stage the efficiency of the perceptual GAN was tested. Perceptual GAN was trained on data from unbalanced original dataset and data generated by DCGAN. These approaches were tested on VOC Pascal dataset with Faster R-CNN model. The VOC Pascal dataset consists mainly of big objects, which enlarge the small object accuracy disparity, as the model focuses mainly on medium and big objects. Moreover, there is a significant disproportion in class count depending on object size, which results in a lack of diversity and location of small objects.
The experiments face three problematic aspects of small object detection: the low number of samples, their lack of diversity, and low features representation. The first phase involves dataset preparation by augmenting classes for small objects from the VOC Pascal dataset with several oversampling strategies. The original objects used for the oversampling method are enhanced by the ones generated by Generative Adversarial Network, based on the original paper \cite{dcgan} customized for the training purpose. The augmented dataset is introduced to the Faster R-CNN model and evaluated on original ground-truth images. Secondly, for a selected class from training augmented dataset with low classification results, FGSM attack is conducted on objects as a trial to increase the identification score. Finally, in order to cope with poor small object representation, the enhanced dataset is introduced to Perceptual GAN, which generates super-resolved large-object like representations for small objects and enables higher recognizability.

\section{Related works}
To address the small object detection problem numerous methods are introduced with different results. In order to cover the small object accuracy gap between one and two-stage detectors, Focal Loss \cite{focal-loss} can successfully be applied. It involves a loss function modification that puts more emphasis on misclassified examples. The contribution of correctly learned, easy examples is diminished during the training with the focus on difficult ones. With high-resolution photos and relatively small objects, the detection accuracy can also be improved by splitting the input image into tiles, with every tile separately fed into the original network. 
The so-called Pyramidal feature hierarchy \cite{feature-pyramid-hierarchy} addresses the problem of scale-invariant object detection. By replacing the standard feature extractor, it allows creating better quality, high-level multi-scale feature maps. The mechanism involves two inverse pathways. The feature maps computed in the forward pass are upsampled to match the previous layer dimension and added element-wise. In this way, the abstract low-level layers are enhanced with higher-level semantically stronger features the network calculated close its head, which facilitates the detector's small objects pick up. The evaluations on the MS COCO dataset allowed increasing the overall mAP from 47.3 up to 56.9.

Some other approach \cite{finding-tiny-faces} faces problem subdomain - face detection, tries to make use of object context. The detectors are trained for different scales on features extracted from multiple layers of feature hierarchy. Using context information to improve small object accuracy is also applied in \cite{small-object-detection-context-attention}. In this work, the authors firstly extract object context from surrounding pixels by using more abstract features from high-level layers. The features of the object and the context are concatenated providing enhanced object representation. The evaluation is performed on SSD with attention module (A-SSD) to allow the network to focus on important parts rather than the whole image. Comparing with conventional SSD the method achieved significant enhancement for small objects from 20.7\% to 28.5\%.
Another modification of the Feature Proposal Network (FPN) approach applied to Faster RCNN \cite{small-object-detection-multiscale-features} extracts features of the 3rd, 4th, and 5th convolution layers for objects and uses multiscale features to boost small object detection performance. The features from the higher levels are concatenated with the ones from the lowers into a single dimension vector, running a 1x1 convolution on the result. This allowed a 0.1 increase in the mAP in regards to the original Faster R-CNN model.

The small number of samples issue is faced in \cite{augmentation-small-object-detection}, where the authors use the oversampling method as a small object dataset augmentation technique and reuse the original object to copy-paste it several times. In this way, the model is encouraged to focus on small objects, the number of matched anchors increases, which results in the higher contribution of small objects to the loss function. Evaluated on Mask R-CNN using MS COCO dataset, the small objects AP increased while preserving the same performance on other object groups. The best performance gain is achieved with oversampling ratio equal to 3.
Some generative models \cite{srgan-explain} attempt to achieve super-resolution representations for small objects, and in this way facilitate their detection. Those frameworks already have capabilities of inferring photo-realistic natural images for 4x upscaling factors, however, they require heavy time consumption for training. The proposed solution uses a deep residual network (ResNet) in order to recover downsampled images. The model loss includes an adversarial loss, that pushes discriminator to make a distinction between super-resolution images and original ones, and content loss to achieve perceptual similarity instead of pixel space similarity. SRGAN derivative - classification-oriented SRGAN \cite{csr-gan} append classification branch and introduce classification loss to typical SRGAN, generator of CSRGAN is trained to reconstruct realistic super-resolved images with classification-oriented discriminative features from low-resolution images while discriminator is trained to predict true categories and distinguish generated SR images from original ones. Some other approach \cite{improve-object-detection-data-enhancement} proposes a data augmentation based on the foreground-background segregation model. It adds an assisting GAN network to the original SSD training process. The first training phase focus on the foreground-background model and pre-training object detection. The second stage encloses a certain probability data enhancement such as color channel change, noise addition, and contrast boost. The proposed method increases the overall mAP to 78.7\% (SSD300 baseline equals 77.5\%). Another Super-Resolution Network SOD-MTGAN \cite{sod-mt-gan} aims to create the images where it will be easier for the resulting detector, which is trained along the side of the generator, to actually locate the small objects. So, the generator here is used to upsize blurred images to better quality and create descriptive features for those small objects. The discriminator, apart from differentiating between real and generated images, describes them with category score and bounding box location. 

The Perceptual GAN presented in \cite{PCGAN} has the same goal as the previous super-resolution network but slightly different implementation. Its generator learns to transfigure poor representations of the small objects to super-resolved ones that are commensurate to real large objects to deceive a competing discriminator. Meanwhile, its discriminator contends with the generator to identify the generated representation and enforces an additional perceptual loss – generated super-resolution representations of small objects must be useful for the detection task.
The small objects problem has been already noticed with some enhancement methods proposed, however, in this area there is still much room for improvement, as often described methods are domain-specific and apply to certain datasets. Generative Adversarial Networks are worth further exploration in the object detection area.

\section{Dataset analysis}


Three size groups are extracted from the VOC Pascal dataset according to the annotation bounding boxes: small (size below 32x32), medium (size between 32x32 and 64x64), and big (size above 64x64). Corresponding XML annotations are saved per each category containing only objects with selected size. Tables 1 and 2 
present object distribution in regards to the category and size, with significant variances. For the trainval dataset, small objects constitute less than 6\% of the total objects count. Excluding difficult examples, this number reduces to 1.3\%. Similar statistics apply to the test dataset. Those numbers confirm the problem described earlier. There is a significant disproportion in the object numerosity for different size groups.
The categorized VOC Pascal dataset is introduced to the pre-trained PyTorch Faster R-CNN model described above with the accuracy metrics presented in Table 3 
Overall the network’s performance on small objects (3.13\%) is more than 20 times worse than on big objects (70.38\%). Additionally, the number of samples per category differs substantially, which at least partially concurs to the very low accuracy scores. Only 5.9\% of annotated objects from trainval dataset belong to small objects, whereas medium and big objects take 15.36\% and 78.74\% respectively.
The low number of samples and poor representation of smaller objects is one of the major obstacles in further work, as it disables the network to learn the right representation for the object detection network. The great disparity between big and small objects count bias the Faster R-CNN training to focus on bigger objects. Moreover, as shown in Table 3, 
 there is a significant disproportion in the class numerosity. Basing on the publicly available DCGAN model, a customized, stable GAN implementation is introduced in order to increase the variety of small objects and provide clearer representation. The selected solution includes objects augmentation instead of whole images. 

\section{Data augmentation with DCGAN}
One of the augmentation technique used in experiments was
small object generation with DCGAN (deep convolutional GAN) \cite{radford}. The model’s discriminator is made up of convolution layers, batch norm layers, and leaky ReLU activations. The discriminator input is a 3x32x32 image and the network’s output is a scalar probability that the input is from the real data distribution. The generator is comprised of a series of convolutional-transpose layers, batch norm layers, and ReLU activations. The input is a 100-dimensional latent vector, z, extracted from a standard Gaussian distribution and the output is a 3x32x32 RGB image. The initial model weights are initialized randomly with a normal distribution with mean 0 and stdev 0.02. Both models use Adam optimizers with learning rate 0.0002 and beta = 0.5. The batch size is 64. Additionally, in order to improve the network’s stability and performance, some adjustments are introduced. The training is split into two parts for the generator and the discriminator, as different batches for real and fake objects are constructed. Secondly, to equalize the generator and discriminator training progress, soft and noisy data labels are introduced. Instead of labeling real and fake data as 1 and 0, a random number from range 0.8 - 1.0 and 0.0 - 0.2 is chosen. Moreover, the generator uses dropouts after each layer (25\%). The generator’s progress is assessed manually, by generating a fixed batch of latent vectors that are drawn from a Gaussian distribution and periodically input to the generator. The evaluation includes both the quality and diversity of the images in relation to the target domain. The typical training lasts from 1000-2000 epochs depending on dataset numerosity. In Figure 1 the bird class generation is presented.


\begin{center}
\begin{table}[h]
\centering
\begin{tabular}{ |p{4cm}|c|c|c|c|c|c|c|c|c|c| }
\hline
type & airplane & bike & bird & boat & bottle & bus & car & cat & chair & cow \\
\hline
test small & 25 & 12 & 57 & 65 & 40 & 6 & 149 & 1 & 58 & 28 \\
test small (non diff) & 23 & 0 & 12 & 4 & 1 & 1 & 33 & 1 & 0 & 3 \\
trainval small & 19 & 15 & 36 & 25 & 62 & 12 & 173 & 0 & 34 & 21 \\
trainval small (non diff) & 15 & 1 & 8 & 1 & 10 & 1 & 35 & 0 & 2 & 1 \\
\hline
test medium & 35 & 31 & 116 & 87 & 186 & 18 & 339 & 6 & 250 & 96 \\
trainval medium & 39 & 36 & 119 & 90 & 168 & 18 & 398 & 9 & 258 & 56 \\
\hline
test big & 251 & 346 & 403 & 241 & 431 & 230 & 1053 & 363 & 1066 & 205 \\
trainval big & 273 & 367 & 444 & 283 & 404 & 242 & 1073 & 380 & 1140 & 279 \\
\hline
\end{tabular}

\caption{Object number statistics for classes from VOC pascal (airplane, bike, bird, boat, bottle, bus, car, cat, chair, cow) for trainval and test set with the division for small, medium, big categories.}
\end{table}
\label{tab:voc_analysis_1}
\end{center}

\begin{center}
\begin{table}[h]
\centering
\begin{tabular}{ |p{4cm}|c|c|c|c|c|c|c|c|c|c| }
\hline
type & table & dog & horse & moto & person & plant & sheep & sofa & train & tv \\
\hline
test small & 1 & 4 & 4 & 9 & 305 & 35 & 47 & 0 & 0 & 15 \\
test small (non diff) & 0 & 0 & 1 & 1 & 99 & 15 & 14 & 0 & 0 & 7 \\
trainval small & 0 & 0 & 3 & 11 & 370 & 41 & 87 & 0 & 1 & 14 \\
trainval small (non diff) & 0 & 0 & 1 & 1 & 89 & 20 & 22 & 0 & 0 & 2 \\
\hline
test medium & 8 & 18 & 34 & 36 & 811 & 123 & 57 & 0 & 11 & 64 \\
trainval medium & 11 & 18 & 15 & 30 & 819 & 159 & 79 & 6 & 13 & 65 \\
\hline
test big & 290 & 508 & 357 & 324 & 4111 & 434 & 207 & 396 & 291 & 282 \\
trainval big & 299 & 520 & 388 & 349 & 4258 & 425 & 187 & 419 & 314 & 288 \\
\hline
\end{tabular}
\caption{Object number statistics for classes from VOC pascal (table, dog, horse, motorbike, person, potted plant, sheep, sofa, train, tv monitor) for trainval and test set with the division for small, medium, big categories.}
\end{table}
\label{tab:voc_analysis_2}
\end{center}

\begin{center}
\begin{table}[h]
\centering
\begin{tabular}{ |c|c|c|c|c| }
\hline
dataset & all objects & small objects & medium objects & big objects \\
\hline
mAP & 69.98 & 3.13 & 12.62 & 70.38 \\
\hline
number of images (test) & 4 952 & 366 & 995 & 4843 \\
number of objects (test) & 14 976 & 861 & 2 326 & 11 789 \\
\hline
number of images (trainval) & 5 011 & 378 & 486 & 4 624 \\
number of objects (trainval) & 15 662 & 924 & 2 406 & 12 332 \\
\hline
\end{tabular}
\label{tab:voc_analysis}
\caption{Mean average precision metric in percent for VOC Pascal small, medium, big objects for pre-trained Faster R-CNN model. Below the number of samples per size category for trainval and test images (both include objects marked as difficult). As one image may contain objects from multiple size groups, it’s ID may find in different size categories. That the reason why the number of small, medium and big object images does not bring the cumulative number of images.}
\end{table}
\end{center}

\begin{center}
    \begin{figure}[H]
    \centering
      \includegraphics[scale=0.45]{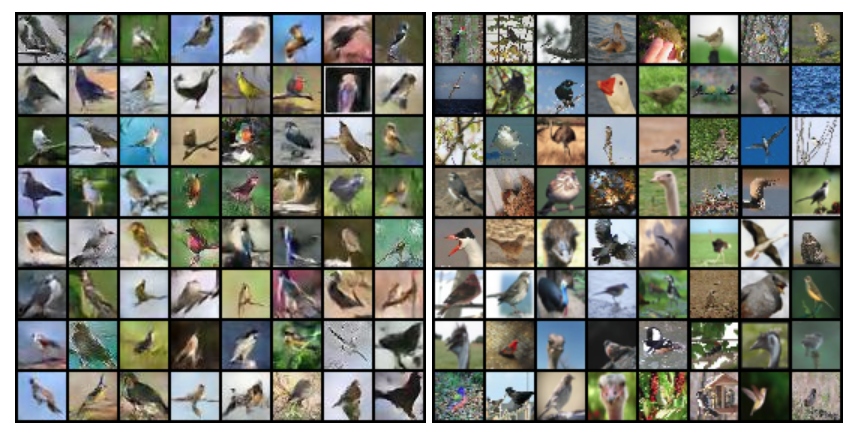}
        \caption{DCGAN generated samples of a bird (left), original training dataset for the bird category (right).}
    \end{figure}
\end{center}


GAN training requires a considerable amount of data of low dimensionality and clear representation, which results in generated objects quality. The main evaluation object detection dataset in presented work is VOC Pascal, however taking into account that it is a relatively small dataset, despite including into training all size groups for a given object. To make training of DCGAN more efficient the dataset has to be enhanced by some other images. The samples from following multiple datasets were tried for the training: Stanford-cars, CINIC-10, FGVC-aircraft, MS COCO, CIFAR-100, 102 Category Flower Dataset, ImageNet and Caltech-UCSD. The seven categories were taken as a case study: car, airplane, tv monitor, boat, potted plant, bird and horse. Table 4 presents the datasets that are successfully applied for object generation with the corresponding count. Having to fulfill the input requirements some data preprocessing procedures had to be conducted in order to obtain a 32x32 RGB image. 

Despite the numerosity, ImageNet dataset training does not bring positive results. The representation of many images are not clear enough, multiple objects are present on single category image and downsampling high-resolution images outputs noise. Similar results are obtained for the MS COCO dataset, with the majority of rectangle objects, after downsampling to square 32x32 resolution, the preprocessed images present noise. In conclusion, a successful dataset for deep GAN should consist of at least several thousands of samples of low resolution, clear representation images per each generated class. 

\begin{center}
\begin{table}[h]
\centering
\begin{tabular}{ |p{2.5cm}|p{10.5cm}|p{2cm}| }
\hline
Category & Datasets & Count \\
\hline
car & Stanford-cars, CINIC-10, VOC Pascal & 22 928 \\
\hline
airplane & FGVC-aircraft, CINIC-10, MS COCO, VOC Pascal & 19 358 \\
\hline
tv monitor & MS COCO, VOC Pascal & 10 175 \\
\hline
boat & MS COCO, CINIC-10, VOC Pascal & 28 793 \\
\hline
potted plant & CIFAR-100, ImageNet, MS COCO, 102 Category Flower Dataset, VOC Pascal & 17 772 \\
\hline
bird & Caltech-UCSD, CINIC-10, VOC Pascal & 12 363 \\
\hline
horse & CINIC-10, MS COCO, VOC Pascal & 33 012 \\
\hline
\end{tabular}
\caption{Datasets used for DCGAN training with the cumulative number of samples per each used category.}
\end{table}
\end{center}

\section{Augmentation setup}

\subsection{Oversampling strategies}

The training dataset for learning FasterRCNN is augmented with several oversampling techniques. As a common rule, the objects are copied from the original location and pasted to a different position, which does not overlap with other objects and the image boundaries. The oversampling ratio equals three. This easy method allows enlarging the area covered by small objects and puts more emphasis on smaller object loss during the training stage. The experiment’s overview is summarized in Table 5. There are two sets of experiments conducted. In the first one, the VOC Pascal objects are used as oversampling strategy in three different scenarios. First, the original small object is picked per image and copied-pasted 3 times in random locations keeping the original object size (strategy 1). Second, instead of multiplying the original object, a random VOC Pascal object for the matching category is picked with returns for every copy-paste action. Each oversampled object is rescaled to a random width and height taken from a range of current width and height and 32 pixels (strategy 2). The oversampling increases the objects count, however taking into account, that the original number of small object samples is around 30 times less than the bigger ones, this oversampling strategy would result in the dataset size increased by 3, still leaving the considerable count disproportion. To address this problem, every image containing a small object is used 5 times with the described random oversampling strategy, which results in multiple original images oversampled with different objects for a given category (strategy 3). The third strategy additionally involves object class modification. The previous method is extended with the following assumption, for every picked small object a random VOC Pascal category is chosen, from the set of most numerous small test categories. This excludes the following classes: sofa, dog, table, and cat for which the number of both train and test samples is below 5. 
The second set of experiments makes use of DCGAN generated objects. 
The generated objects are switched for the following categories: airplane, bird, boat, car, chair, horse, person, potted plant, and tv monitor. There are two test settings similar to the ones conducted for original VOC Pascal objects augmentation. In the first set, for every original small object, a random object of the same category is picked. For DCGAN available classes, generated objects are used, for the others (bike, bottle, bus, cow, motorbike, sheep) the samples come from VOC Pascal. Similarly to the first experiment set, every image containing a small object is used for oversampling 5 times with the oversampling ratio equals 3. The second setting preserves the conditions from the first one and additionally switches the class for every object, meanwhile increasing the oversampling strategy count. For less numerous classes with trainval count below 15 (airplane, train, bicycle, horse, motorbike, bus), the oversampling procedure is repeated 15 times, for other categories, 10 times. The detailed information about the augmented dataset numerosity for each oversampling strategy is presented in Table 6.
Taking into account the datasets introduced, it is clear that some of them (strategies 1, 2 and 4) present a significant disparity in the number of samples for different categories, as they do not include the class change. In Strategy 2, person class numerosity is more than 100 times of the count for train category. To cover this imparity, the random class modification is introduced in Strategies 3 and 5, which results in more even class distributed dataset. For every experiment, VOC Pascal image file annotations are created with the corresponding objects, while erasing the original annotations in order to avoid duplicate annotations.
The five resulting augmented datasets are introduced to the Faster R-CNN network separately, for each the training is combined with the original trainval VOC Pascal dataset. The results are evaluated on the original ground-truth dataset, divided into three size categories (small, medium and big).

\begin{center}
\begin{table}[h]
\centering
\begin{tabular}{ |p{2cm}|p{13.5cm}| }
\hline
Strategy 1 & Oversample x3 original VOC Pascal object with random size and random location \\
\hline
Strategy 2 & Oversample x3 random VOC Pascal object of the same category with random size and location, the procedure is repeated 5 times for every small object \\
\hline
Strategy 3 & Oversample x3 random VOC Pascal object of the randomly changed category with random size and location, the procedure is repeated 5 times for every small object \\
\hline
Strategy 4 & Oversample x3, for selected classes DCGAN generated objects are used, for others random VOC Pascal objects with random size and location, the object category is preserved, the procedure is repeated 5 times for every small object \\
\hline
Strategy 5 & Oversample x3, for selected classes DCGAN generated objects are used, for others random VOC Pascal objects with random size and location, the object category is randomly chosen, the procedure is repeated 15 times for less numerous categories (aeroplane, train, bicycle, horse, motorbike, bus), 10 times for the others. \\
\hline
\end{tabular}
\caption{Augmentation strategies overview.}
\end{table}
\end{center}

\begin{center}
\begin{table}[h]
\centering
\begin{tabular}{ |c|c|c|c|c|c| }
\hline
 & Strategy 1 & Strategy 2 & Strategy 3 & Strategy 4 & Strategy 5 \\
\hline
all & 3 696 & 6 810 & 6 810 & 6 810 & 12 624 \\
\hline
airplane & 76 & 187 & 381 & 187 & 790 \\
bike & 60 & 48 & 296 & 48 & 703 \\
bird & 144 & 216 & 406 & 216 & 783 \\
boat & 100 & 235 & 413 & 235 & 762 \\
bottle & 248 & 422 & 414 & 422 & 829 \\
bus & 48 & 120 & 426 & 120 & 763 \\
car & 692 & 1 448 & 540 & 1 448 & 858 \\
chair & 136 & 319 & 416 & 319 & 745 \\
cow & 84 & 186 & 367 & 186 & 729 \\
horse & 12 & 51 & 334 & 51 & 742 \\
motorbike & 44 & 116 & 366 & 116 & 699 \\
person & 1 480 & 2 680 & 722 & 2 680 & 1 098 \\
plant & 164 & 251 & 412 & 251 & 728 \\
sheep & 348 & 342 & 457 & 342 & 882 \\
train & 4 & 25 & 381 & 25 & 712 \\
tv & 56 & 164 & 379 & 164 & 801 \\
\hline
\end{tabular}
\caption{Augmented dataset count summary, distributed per classes and augmentation strategies. Strategies 1, 2 and 4, as they do not include the class change, present a significant disparity in the number of samples for different categories. In Strategy 2, person class numerosity is more than 100 times of the count for train category. To cover this imparity, the random class modification is introduced in Strategies 3 and 5, which results in more even class distribution.}
\end{table}
\end{center}

\subsection{Perceptual GAN}

Having prepared the equally size distributed dataset and ensured that the generated objects are correctly recognized by the network, they are introduced to Perceptual GAN, which addresses the next reason of poor small object detection - their low feature representation. The PCGAN aims to generate super-resolved large-object like representation for small objects. The approach is similar to architecture described in paper \cite{PCGAN}. The generator model is a modified Faster R-CNN network. 
The generator network is based on Faster R-CNN with residual branch which accepts the features from the lower-level convolutional layer (first conv layer) and passes them to the 3x3 and 1x1 convolutions, followed by max pool layer. As a next step there are two residual blocks with the layout consisting of two 3x3 convolutions, batch normalizations with ReLU activations which aim to learn the residual representation of small objects. The super-resolved representation is acquired by the element-wise sum of the learned residual representation and the features pooled from the fifth conv layer in the main branch.

The learning objective for vanilla GAN models \cite{goodfellowgan} corresponds to a minimax two-player game, which is formulated as (eq. \ref{gan}):
\begin{equation}
\minmax_{G \ \ \ D} \ \ L(D,G) = E_{x \sim p_{data(x)}} log D(x) + E_{z \sim p_{data(z)}}[log(1-D(G(z)))]
\label{gan}
\end{equation}

G represents a generator that learns to map data $z$ (with the noise distribution $p_{z}(z)$) to the distribution $p_{data}(x)$ over data $x$. D represents a discriminator that estimates the probability of a sample coming from the data distribution $p_{data}(x)$ rather than $p_{z}(z)$. The training procedure for G is to maximize the probability of D making a mistake.

The $x$ and $z$ are the representations for both large objects and small objects, i.e., $F_l$ and $F_s$ respectively. The goal is to learn a generator function which transforms the representations of a small object $F_s$ to a super-resolved one
G($F_s$) that is similar to the original one of the large object $F_l$. 
Therefore, a new conditional generator model is introduced which is conditioned on the extra auxiliary information, i.e., the low level features of the small object f from which the generator learns to generate the residual representation between the representations of large and small objects through residual learning (eq.\ref{residual}).

\begin{equation}
\minmax_{G \ \ \ D} \ \ L(D,G) = E_{F_l \sim p_{data(F_l)}} log D(F_l) + E_{F_s \sim p_{F_s}}[log(1-D(F_s+G(F_s|f)))]
\label{residual}
\end{equation}

In this case, the generator training can be substantially simplified over directly learning the super-resolved representations for small objects. For example, if the input representation is from a large object, the generator only needs to learn
a zero-mapping. 

The original’s paper discriminator consists of two branches \cite{PCGAN}: adversarial to distinguish generated super resolved representation from the original one for the large object and perception, to validate the accuracy influence of generated super-resolved features. In this solution, the perception branch is omitted with the main emphasis put on an adversarial branch. The adversarial branch consists of three fully connected layers, followed by sigmoid activation producing an adversarial loss. For the training purpose, there are two datasets prepared, containing images with only small and big objects respectively. The images are resized to 1000 x 600 pixels. To solve the adversarial min-max problem the parameters in the generator and the discriminator networks are optimized. Denote $G_{\Theta_g}$ 
as the generator network with parameters
$\Theta_g$. The $\Theta_g$ is obtained by optimizing the loss function $L_{dis}$ ($L_{dis}$ is the adversarial loss, eq.\ref{generator}).



\begin{equation}
\Theta_g = \argmin_{\ \ \Theta_g} \ \ L_{dis}(G_{\Theta_g}(F_s)) 
\label{generator}
\end{equation}

Suppose $D_{\Theta_a}$ is the adversarial branch of the discriminator network parameterized by $\Theta_a$. The $\Theta_a$ is obtained by optimizing a specific loss function $L_a$ (eq.\ref{adversarial}).

\begin{equation}
\Theta_a = \argmin_{\Theta_a} \ \ L_{a}(G_{\Theta_G}(F_s), F_l) 
\label{adversarial}
\end{equation}

The loss $L_a$ is defined as:

\begin{equation}
L_a = -log D_{\Theta_a}(F_l) + log(1-D_{\Theta_a}(F_s+G_{\Theta_G}(F_s)))]
\end{equation}

The $L_a$ loss encourages the discriminator network to distinguish the difference between the currently generated super-resolved representation for the small object and the original one from the real large object.


In the first phase, the generator is fed with large objects with the real batch forward pass through the discriminator. Next, the generator is trained with the small object dataset, trying to maximize the loss $log(D(G(z))$, where $G(z)$ is the fake super-resolved small object image. The generator’s loss, apart from the adversarial loss justifying the probability of the input belonging to a large object, acknowledges the RPN and ROI losses. The whole network is trained with Stochastic Gradient Descent with the momentum 0.9 and learning rate 0.0005. 
The perceptual GAN training is performed separately for two oversampled datasets representing the small object dataset and original VOC Pascal trainval set for large objects. For the small object dataset, the oversampling strategy with VOC Pascal objects combined with the random class switch was used. First, the evaluation is conducted on VOC small objects subset. Then ensemble model is created with original Faster-RCNN and PCGAN and voting mechanism is added at the end. This solution allows to detect large objects at the similar level as before and increase detection accuracy of small objects.

\section{Results}

Table 7 
shows the mAP score achieved by the FasterRCNN model trained with datasets obtained with described augmentation strategies, evaluated on the original VOC Pascal dataset.

\begin{center}
\begin{table}[h]
\centering
\begin{tabular}{ |c|c|c|c|c| }
\hline
Strategy & small obj count & mAP - small & mAP - medium & mAP - big \\
\hline
original & 861 & 3.10 & 12.62 & 70.38 \\
\hline
Strategy 1 & 3696 & 5.79 & 12.71 & 66.80 \\ 
Strategy 2 & 6810 & 5.84 & 12.95 & 67.75 \\
Strategy 3 & 6810 & 7.08 & 14.17 & 66.73 \\
Strategy 4 & 6810 & 5.47 & 13.56 & 67.84 \\
Strategy 5 & 12 624 & 7.60 & 16.28 & 67.01 \\
\hline
\end{tabular}
\caption{Evaluation of different augmentation strategies, described in section 5. The tests are conducted on VOC Pascal dataset, splitted into three size categories (see section 5.1). Mean Average Precision is given as percentage value with additional information about the augmented small object train dataset count.}
\end{table}
\label{tab:strategies_}
\end{center}

The strategies including random class modification for oversampling with VOC Pascal and generated objects (strategies 3 and 5) outperform the original results by 3.98\% and 4.5\% respectively. Generally, by increasing the number of samples during the training, the mAP on small objects can be improved without any model modification. As proved, even the most naive solution, by oversampling the original object without any changes allowed to achieve almost two times better score. The most gain is observed with oversampling using DCGAN generated objects. However, the accuracy differences between using VOC Pascal and generated objects are quite low ($\sim$0.6\%). In addition, augmenting small objects affected medium objects’ performance. In case of the first two strategies, mAP remained unchanged, for the other cases it achieved a better score than the original. The best performance is registered for the last oversampling strategy, which assumed augmentation with generated objects together with random class switch with a score 16.28\%. It outperforms original results and augmented by VOC objects by 3.66\% and 2.11\% respectively. Summing up, the strategy 5 oversampling method produced the overall best results. Firstly, due to the highest number of samples used distributed evenly between categories, secondly by enhancing the representation of the objects.

In order to demonstrate where does the improvement come from, FasterRCNN results over classes are presented. Tables 8 and 9 show the results of oversampling scenarios evaluated on test VOC Pascal dataset splitted by categories. Overall, the presented augmentation strategies bring improvement to most analyzed categories with oversampling performed. This applies to the following classes: airplane, bird, boat, bus, car, cow, horse, motorcycle, person, potted plant, sheep and tv monitor. As explained in section 4 some categories are not the subject to augmentation procedure due to very low number of original train and test samples (sofa, dog, table and cat). For the remaining classes: bike, bottle, chair, despite boosting the trainval set representation, there is no detection improvement observed. The reason might be the quality and the features of the train and test subset for those categories, where the majority of small objects is classified as difficult. This outcome is also confirmed by the chair class, where applying representative DCGAN generated objects, once again does not influence the achieved mAP score as all 58 test objects are difficult ones. Additionally, it is worth mentioning that the context plays a significant role for the airplane category. The best results are obtained for the second strategy with no class change. 
Another interesting observation is the fact that the bird and cow are the categories that benefits most from DCGAN generated objects. The mAP score for bird is 5.34\%, which is 4.45\% and 1.6\% better than the original and best VOC oversampling strategy, respectively. The cow category has at least two times better mAP score using DCGAN based oversampling. On the other hand the airplane is the only category, for which the generated objects used in strategy 5 do not improve the detection accuracy. 

\begin{center}
\begin{table}[h]
\centering
\begin{tabular}{ |p{4cm}|c|c|c|c|c|c|c|c|c|c| }
\hline
type & airplane & bike & bird & boat & bottle & bus & car & chair & cow \\
\hline
orginal & 6.09 & 0.00 & 0.89 & 1.19 &  0.00 & 0.00 & 2.19 & 0.00 & 3.16 \\
Strategy 1 & 7.04 & 0.00 & 0.69 & 1.09 & 0.00 & 0.00 & 4.31 & 0.00 & 10.98 \\
Strategy 2 & 18.70 & 0.00 & 2.04 & 0.00 & 0.00 & 0.83 & 4.26 & 0.00 & 4.16 \\
Strategy 3 & 10.72 & 0.00 & 3.74 & 0.85 & 0.00 & 2.38 & 4.38 & 0.00 & 7.08 \\
Strategy 4 & 2.42 & 0.00 & 5.34 & 1.47 & 0.00 & 0.00 & 4.39 & 0.00 & 22.52 \\
Strategy 5 & 5.37 & 0.00 & 1.66 & 1.36 & 0.00 & 0.00 & 3.99 & 0.00 & 41.52 \\
\hline
\end{tabular}
\caption{VOC Pascal categories mAP scores (airplane, bike, bird, boat, bottle, bus, car, chair, cow) for trainval and test set with the division for small, medium, big categories. Mean Average Precision is given as percentage value}
\end{table}
\end{center}

\begin{center}
\begin{table}[h]
\centering
\begin{tabular}{ |p{4cm}|c|c|c|c|c|c|c|c|c|c| }
\hline
type & horse & moto & person & plant & sheep & tv \\
\hline
original & 2.63 & 0.00 & 0.92 & 0.44 & 20.07 & 6.20 \\
Strategy 1 & 2.56 & 3.57 & 1.24 & 3.93 & 29.09 & 16.34 \\
Strategy 2 & 3.03 & 4.00 & 1.28 & 5.17 & 25.29 & 13.07 \\
Strategy 3 & 7.14 & 5.00 & 1.43 & 4.75 & 34.16 & 17.47 \\
Strategy 4 & 1.96 & 2.17 & 1.32 & 2.31 & 27.39 & 5.23 \\
Strategy 5 & 3.03 & 5.55 & 1.17 & 1.03 & 25.66 & 16.43 \\
\hline
\end{tabular}
\caption{VOC Pascal categories mAP scores (horse, motorbike, person, potted plant, sheep and tv monitor) for trainval and test set with the division for small, medium, big categories. Mean Average Precision is given as percentage value}
\end{table}
\end{center}

Table 10 provides the summary of PCGAN training results on augmented VOC Pascal. For this process, the third oversampling strategy dataset is used as a small object image dataset, instead of original VOC Pascal, in order to obtain similar number of small and big objects, required for the training phase. Overall, PCGAN allowed increasing mAP for small objects in nine presented classes. It is apparent, that the solution allows much better performance than simple augmentation strategies. The most gain in the mAP score is observed for the motorcycle class, from the original 0\%, through 5\% (oversampling) up to 50\% (perceptual GAN). The other significant improvement represent bird category with a score 11.62\%, which is 2.3 and 13.1 times better than oversampling and original result. For aeroplane, boat, bus, cow, sheep and tv monitor categories the mAP performance fluctuates around 1.5 times better than oversampling score. Worth mentioning is a cow class, for which FasterRCNN achieved over 65\% accuracy.
The following observation may be extracted at this point. Firstly, the Perceptual GAN can be successfully extended to natural scene image dataset from its initial application. Secondly, the dedicated solution, such as PCGAN, despite heavier training procedure (generator and discriminator), allows to achieve significantly better detection accuracy for small objects than augmentation methods. The described solution may be efficiently employed with the original FasterRCNN as a parallel network and forms ensemble model. The second possibility is to use conditional generator as described in section 5.2. In both solutions the original mAP score for big objects is preserved. The small object mAP is significantly improved. The score for medium object is at the same level or slightly better than in original model. After applying these approaches mAP for whole dataset is improved up to 0.3\% (from 69.98\% up to $\sim$70.3\%). The small increase is dictated by small percentage of small and medium objects in test dataset (Table 1 and 2). 


\begin{center}
\begin{table}[h]
\centering
\begin{tabular}{ |c|c|c|c|c|c|c|c|c|c| }
\hline
Strategy & aeroplane & bird & boat & bus & cow & horse & moto & sheep & tv \\
\hline
original & 6.09 & 0.89 & 1.20 & 0.00 & 3.16 & 2.63 & 0.00 & 20.07 & 6.20 \\
best oversampling & 18.70 & 5.34 & 1.47 & 2.38 & 41.52 & 7.14 & 5.55 & 34.16 & 17.47 \\
\hline
PCGAN & 28.97 & 11.62 & 2.49 & 3.45 & 65.19 & 9.09 & 50.0 & 44.58 & 27.27 \\
\hline
\end{tabular}
\label{tab:pcgan}
\caption{Evaluation of PCGAN training described in 5.2. The results are presented in comparison with original FasterRCNN and oversampling strategy that received best score for given category. All three tests are conducted on small object VOC Pascal test group. Mean Average Precision is given as percentage value.}
\end{table}
\end{center}

For all simulation presented in the paper the ratios of width and height of the generated FasterRCNN anchors used are 0.5, 1 and 2. The areas of anchors (anchor scales) are defined as 8, 16 and 32 with the feature stride equal 16. The learning rate is 0.01.

\section{Conclusions and future work}

The work presents comparison of few strategies for improving small object detection. The presented results show that solution with GAN architecture outperforms other well known augmentation approaches. The perceptual GAN is significantly better than oversampling strategies based on DCGAN image generation. It achieves better results with the similar amount of the training data. It is worth noting that all presented approaches required a 10-20 fold increase in the number of small objects. Future work will concentrate on further improvements using perceptual GAN. The experiments will focus on perceptual GAN architecture exploration. Next, the solution will be tested on other object detection datasets.

\bibliographystyle{unsrt}  



\begin{thebibliography}{1}

\bibitem{PCGAN}
  Jianan Li, Xiaodan Liang, Yunchao Wei, Tingfa Xu, Jiashi Feng, Shuicheng Yan, Perceptual Generative Adversarial Networks for Small Object Detection, 2017
  {\em https://arxiv.org/pdf/1706.05274.pdf}, 2017-06-20

\bibitem {fasterrcnn-explain}
  Shaoqing Ren, Kaiming He Ross, Girshick Jian Sun,
  Faster R-CNN: Towards Real-Time Object Detection with Region Proposal Networks, https://papers.nips.cc/paper/5638-faster-r-cnn-towards-real-time-object-detection-with-region-proposal-networks.pdf

\bibitem {GAN}
Goodfellow, Ian and Pouget-Abadie, Jean and Mirza, Mehdi and Xu, Bing and Warde-Farley, David and Ozair, Sherjil and Courville, Aaron and Bengio, Yoshua. Generative Adversarial Nets. Advances in Neural Information Processing Systems, pp. 2672-2680, 2014. http://papers.nips.cc/paper/5423-generative-adversarial-nets.pdf


\bibitem {PCGAN}
Jianan Li, Xiaodan Liang, Yunchao Wei, Tingfa Xu, Jiashi Feng, Shuicheng Yan. Perceptual Generative Adversarial Networks for Small Object Detection, 2017, https://arxiv.org/pdf/1706.05274.pdf.


\bibitem {VOC}
Mark Everingham, Luc van Gool, Chris Williams,
John Winn, Andrew Zisserman. The PASCAL Visual Object Classes Homepage, 2014, http://host.robots.ox.ac.uk/pascal/VOC/
  

\bibitem {rcnn-explain}
Ross Girshick Jeff Donahue Trevor Darrell Jitendra Malik. Rich feature hierarchies for accurate object detection and semantic segmentation, 2014, https://arxiv.org/pdf/1311.2524.pdf

\bibitem {fastrcnn-explain}
Ross Girshick. Microsoft Research. Fast R-CNN, 2017,
https://arxiv.org/pdf/1504.08083.pdf

\bibitem {yolo-explain}
Joseph Redmon, Santosh Divvala, Ross Girshick, Ali Farhadi. You Only Look Once: Unified, Real-Time Object Detection, 2016, https://arxiv.org/pdf/1506.02640.pdf.


\bibitem {ssd-explain}
Wei Liu, Dragomir Anguelov, Dumitru Erhan, Christian Szegedy,Scott Reed, Cheng-Yang Fu , Alexander C. Berg. SSD: Single Shot MultiBox Detector, 2016, https://arxiv.org/pdf/1512.02325.pdf


\bibitem {srgan-explain}
Christian Ledig, Lucas Theis, Ferenc Huszar, Jose Caballero, Andrew Cunningham, Alejandro Acosta, Andrew Aitken, Alykhan Tejani, Johannes Totz, Zehan Wang, Wenzhe Shi Twitter. Photo-Realistic Single Image Super-Resolution Using a Generative Adversarial Network, 2017, https://arxiv.org/pdf/1609.04802.pdf.


\bibitem{cyclegan-explain}
Jun-Yan Zhu, Taesung Park, Phillip Isola Alexei A. Efros. Unpaired Image-to-Image Translation using Cycle-Consistent Adversarial Networks, 2018, https://arxiv.org/pdf/1703.10593.pdf.

\bibitem{dcgan}
Alec Radford,  Luke Metz, Soumith Chintala. Unsupervised Representation Learning with Deep Convolutional Generative Adversarial Networks, 2016, https://arxiv.org/pdf/1511.06434.pdf.

\bibitem{ciraf-website}
Alex Krizhevsky, Vinod Nair, Geoffrey Hinton. Cifar-10, Cifar-100 dataset, 2009, https://www.cs.toronto.edu/~kriz/cifar.html.

\bibitem {vgg}
Karen Simonyan, Andrew Zisserman. Very Deep Convolutional Networks for Large-Scale Image Recognition, 2015, https://arxiv.org/pdf/1409.1556.pdf.

\bibitem {resnet}
Kaiming He, Xiangyu Zhang, Shaoqing Ren, Jian Sun. Deep Residual Learning for Image Recognition, 2015, https://arxiv.org/pdf/1512.03385.pdf.


\bibitem {gan-use-case-medical}
Maayan Frid-Adar, Idit Diamant, Eyal Klang, Michal Amitai, Jacob Goldberger, Hayit Greenspan. GAN-based Synthetic Medical Image Augmentation for increased CNN Performance in Liver Lesion Classification, 2018, https://arxiv.org/pdf/1803.01229.pdf.

\bibitem{gan-use-case-adversarial}
Chaowei Xiao, Bo Li, Jun-Yan Zhu, Warren He, Mingyan Liu, Dawn Song. Generating Adversarial Examples with Adversarial Networks, 2019, https://arxiv.org/pdf/1801.02610.pdf.

\bibitem{gan-use-case-at-gan}
Xiaosen Wang, Kun He, Chuanbiao Song, Liwei Wang, John Hopcroft. AT-GAN: An Adversarial Generator Model for Non-constrained Adversarial Examples, 2020, https://arxiv.org/pdf/1904.07793.pdf.

\bibitem{gan-use-case-unrestricted-adversarial}
Yang Song, Rui Shu, Nate Kushman, Stefano Ermon. Constructing Unrestricted Adversarial Examples with Generative Models, 2018, https://arxiv.org/pdf/1805.07894.pdf.

\bibitem{ms-coco}
Tsung-Yi Lin, Michael Maire, Serge Belongie, Lubomir Bourdev, Ross Girshick, James Hays, Pietro Perona, Deva Ramanan, C. Lawrence Zitnick, Piotr Dollar. Microsoft COCO: Common Objects in Context, 2015 https://arxiv.org/pdf/1405.0312.pdf.


\bibitem{speed-acc-object-detection-comparsion}
Jonathan Huang, Vivek Rathod, Chen Sun, Menglong Zhu, Anoop Korattikara, Alireza Fathi, Ian Fischer, Zbigniew Wojna, Yang Song, Sergio Guadarrama, Kevin Murphy. Speed/accuracy trade-offs for modern convolutional object detectors, 2017, https://arxiv.org/pdf/1611.10012.pdf.


\bibitem{focal-loss}
Tsung-Yi Lin, Priya Goyal, Ross Girshick, Kaiming He, Piotr Dollar. Focal Loss for Dense Object Detection, 2018,
https://arxiv.org/pdf/1708.02002.pdf.


\bibitem{feature-pyramid-hierarchy}
Tsung-Yi Lin, Piotr Dollar, Ross Girshick, Kaiming He, Bharath Hariharan, Serge Belongie. Feature Pyramid Networks for Object Detection, 2017, https://arxiv.org/pdf/1612.03144.pdf


\bibitem{finding-tiny-faces}
Peiyun Hu, Deva Ramanan. Finding Tiny Faces, 2017, https://arxiv.org/pdf/1612.04402.pdf


\bibitem{small-object-detection-context-attention}
Jeong-Seon Lim, Marcella Astrid, Hyun-Jin Yoon, Seung-Ik Lee. Small Object Detection using Context and Attention, 2019, https://arxiv.org/pdf/1912.06319.pdf

\bibitem{small-object-detection-multiscale-features}
Hu, Guo and Yang, Zhong and Hu, Lei and Huang, Li and Han, Jia. Small Object Detection with Multiscale Features, 2018, pp.1-10, International Journal of Digital Multimedia Broadcasting.

\bibitem{augmentation-small-object-detection}
Mate Kisantal, Zbigniew Wojna, Jakub Murawski, Jacek Naruniec, Kyunghyun Cho. Augmentation for small object detection, 2019, https://arxiv.org/pdf/1902.07296.pdf


\bibitem{csr-gan}  
Y. Chen and J. Li and Y. Niu and J. He. Small Object Detection Networks Based on Classification-Oriented Super-Resolution GAN for UAV Aerial Imagery. Chinese Control And Decision Conference (CCDC), 2019, pp.4610-4615.

\bibitem{improve-object-detection-data-enhancement}
Wei Jiang, Na Ying. Improve Object Detection by Data Enhancement based on Generative Adversarial Nets, https://arxiv.org/pdf/1903.01716.pdf


\bibitem{sod-mt-gan}
Yancheng Bai, Yongqiang Zhang, Mingli Ding, and Bernard Ghanem, SOD-MTGAN: Small Object Detection via Multi-Task Generative Adversarial Network, 2018
  
\bibitem{celebA}
Liu, Ziwei and Luo, Ping and Wang, Xiaogang and Tang, Xiaoou, Deep Learning Face Attributes in the Wild, Proceedings of International Conference on Computer Vision (ICCV), December, 2015

\bibitem{stanford-cars}
Jonathan Krause and Michael Stark and Jia Deng and Li Fei-Fei. 3D Object Representations for Fine-Grained Categorization, 4th International IEEE Workshop on  3D Representation and Recognition (3dRR-13) 2013, Sydney, Australia,
  
\bibitem{chair-dataset}
Aubry, Mathieu and Maturana, Daniel and Efros, Alexei, and Russell, Bryan and Sivic, Josef. Seeing 3D chairs: exemplar part-based 2D-3D alignment using a large dataset of CAD models, CVPR, 2014.

\bibitem{goodfellow2014explaining}
Goodfellow, Ian J and Shlens, Jonathon and Szegedy, Christian. Explaining and harnessing adversarial examples,
arXiv preprint arXiv:1412.6572, 2014

\bibitem{goodfellowgan}
I. Goodfellow, J. Pouget-Abadie, M. Mirza, B. Xu, D. Warde-Farley, S. Ozair, A. Courville, and Y. Bengio. Generative adversarial nets. In NIPS, pages 2672–2680, 2014

\bibitem{radford}
Alec Radford, Luke Metz, Soumith Chintala: Unsupervised Representation Learning with Deep Convolutional Generative Adversarial Networks, 2016. URL
https://arxiv.org/pdf/1511.06434.pd

\bibitem{NASNet}
Jonathan Krause and Michael Stark and Jia Deng and Li Fei-Fei Zoph, Barret and Vasudevan, Vijay and Shlens, Jonathon and Le, Quoc. Learning Transferable Architectures for Scalable Image Recognition, pp. 8697-8710, 2018 


\bibitem{inception}
Szegedy, Christian and Liu, Wei and Jia, Yangqing and Sermanet, Pierre and Reed, Scott and Anguelov, Dragomir and Erhan, Dumitru and Vanhoucke, Vincent and Rabinovich, Andrew. Going deeper with convolutions. The IEEE Conference on Computer Vision and Pattern Recognition (CVPR), 2015, pp.1-9

\end{thebibliography}

\end{document}